\documentclass[letterpaper, 10 pt, conference]{ieeeconf}  

\IEEEoverridecommandlockouts                              

\usepackage{cite}
\usepackage{amsmath,amssymb,amsfonts}
\usepackage{algorithmic}
\usepackage{graphicx}
\usepackage{textcomp}
\usepackage{xcolor}

\usepackage{multirow}
\usepackage{wrapfig}
\usepackage{makecell}

\makeatletter
\let\NAT@parse\undefined
\makeatother
\usepackage{hyperref}
\hypersetup{
  colorlinks   = true,    
  urlcolor     = blue,    
  linkcolor    = blue,    
  citecolor    = blue      
}

\usepackage{dsfont}
\usepackage{pifont}

\newcommand{\methodname}{\texttt{MSTA}}
\newcommand{\fullmethodname}{Masked Sensory-Temporal Attention}

\title{\LARGE \bf 
Masked Sensory-Temporal Attention for Sensor Generalization in Quadruped Locomotion
}

\author{Dikai Liu$^{1,2}$ \and Tianwei Zhang$^{2}$ \and Jianxiong Yin$^{1}$ \and Simon See$^{1,3}$
\thanks{$^{1}$ NVIDIA AI Technology Centre (NVAITC); e-mail: {\tt\small \{dikail,jianxiongy,ssee\}@nvidia.com}}%
\thanks{$^{2}$ College of Computing and Data Science, Nanyang Technological University, Singapore; e-mail: {\tt\small dikai001@e.ntu.edu.sg, tianwei.zhang@ntu.edu.sg}}%
\thanks{$^{3}$ also with Nanyang Technological University and Coventry University}%
}

\begin{document}

\maketitle
\thispagestyle{empty}
\pagestyle{empty}

\begin{abstract}
With the rising focus on quadrupeds, a generalized policy capable of handling different robot models and sensor inputs becomes highly beneficial. Although several methods have been proposed to address different morphologies, it remains a challenge for learning-based policies to manage various combinations of proprioceptive information. This paper presents \fullmethodname{} (\methodname{}), a novel transformer-based mechanism with masking for quadruped locomotion. It employs direct sensor-level attention to enhance the sensory-temporal understanding and handle different combinations of sensor data, serving as a foundation for incorporating unseen information. \methodname{} can effectively understand its states even with a large portion of missing information, and is flexible enough to be deployed on physical systems despite the long input sequence.
\end{abstract}

\section{Introduction}
Benefiting from the rapid advancements of deep reinforcement learning (RL) technology~\cite{tan2018sim,hwangbo2019learning, lee2020learning,kumar2021rma}, quadrupedal robots have showcased their capability to navigate in diverse complex terrains. With the increasing availability of affordable quadruped robots on the market, there is a growing interest in developing general-purpose locomotion policies that can fit all types of quadrupedal devices. Unfortunately, existing learning-based locomotion policies are trained for specific models, observation spaces, and tasks, making it challenging to transfer or generalize to other unseen robots or scenarios.

Recently, researchers have developed some generalized policies for quadruped locomotion, such as GenLoco~\cite{feng2023genloco} and ManyQuadrupeds~\cite{shafiee2023manyquadrupeds}, which have the ability to adapt to diverse morphologies. However, these methods still depend on a fixed observation space input for generating latent space representations. They become ineffective when facing the following situations: (1) deployment on quadrupeds with a different sensor set; (2) unreliable sensor data due to wear and tear, (3) adapting to a new task with new input. Since sensory feedback is interrelated and each sensor plays a critical role at different stages of the locomotion~\cite{yu2023identifying}, a policy with a deep understanding of proprioceptive information to handle flexible inputs is desired, to enhance the generalization, flexibility, and extensibility.

\begin{figure}[t]
    \centering
    \includegraphics[width=0.9\columnwidth]{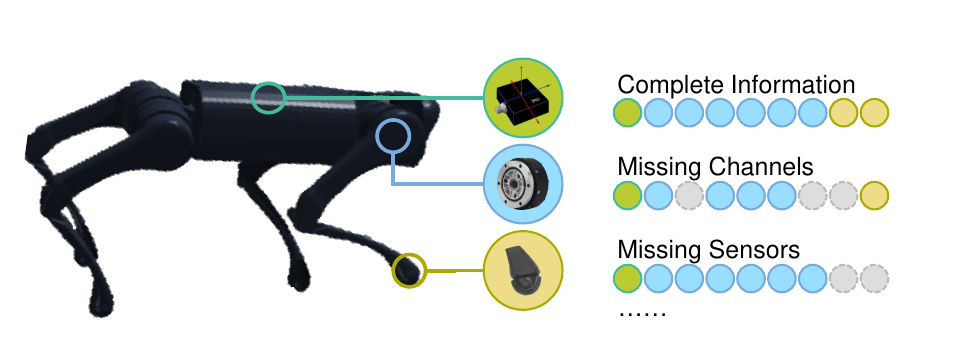}
    \caption{Commonly seen low-level sensors on a quadrupedal robot. However, actual sensor set is still different across models, and sensor degradation can cause part of sensor data to be unreliable or even unavailable. With \methodname{}, we create a generalized model to enhance the understanding of sensor information to handle variable sensor input for quadruped locomotion.}
    \label{fig:sensors}
\end{figure}

One promising solution is self-attention-based transformers~\cite{vaswani2017attention}, which have demonstrated exceptional capabilities in understanding complex sequential information of arbitrary lengths. They have been widely used in robotics to enhance various tasks with multimodal processes~\cite{reed2022generalist, brohan2022rt, brohan2023rt,padalkar2023open}. However, due to the complex model structures and vast parameters, robots driven by transformers often run at very low frequency~\cite{brohan2022rt,brohan2023rt}, or depend on external high-power computing platforms~\cite{fu2024mobile}. For locomotion tasks, the observation-action data are commonly encoded at the timestep level~\cite{caluwaerts2023barkour,lai2023sim,radosavovic2023learning,radosavovic2024humanoid}, which is straightforward and efficient for producing joint commands in an end-to-end manner. However, it limits the transformer's direct access to sensory information and still relies on fixed sensor input as they are hidden behind linear projection, thereby constraining its in-context understanding capability and multimodal nature of the data.

To address the above limitations, we propose \fullmethodname{} (\methodname{}), a novel transformer-based model for end-to-end quadruped locomotion control. It achieves sensor input generalization with its multimodal nature, while still being directly deployable on physical systems. Specifically, in \methodname{}, all sensory data are discretized and tokenized to form a long proprioceptive information sequence. Inspired by the work~\cite{feichtenhofer2022masked} on learning spatiotemporal information in video understanding, a random mask is applied to remove a portion of the observation during training. This significantly enhances the model's sensory-temporal understanding, to better handle different combinations of sensor data and serve as a foundation for incorporating unseen data. Additionally, it aids in identifying the most essential sensory information, thereby reducing the computational power required for physical deployment. 

We conduct extensive experiments in the simulation and physical world. Evaluation results demonstrate that \methodname{} can efficiently handle incomplete sensory information, even with half of the data missing. It is also robust against unseen data, making it a solid foundation for further extensions. With direct sensory-temporal attention, the model is flexible enough to mix-and-match desired information for finetuneing, meeting the requirement for different end-to-end quadruped locomotion control in the physical world. 

\section{Related Work}

\subsection{Sim-to-Real Policy Learning in Legged Locomotion}
Reinforcement learning (RL) has gained significant attention in developing robotic controllers for tasks such as legged locomotion~\cite{tan2018sim,hwangbo2019learning,lee2020learning,kumar2021rma,lai2023sim}, eliminating the need for extensive prior knowledge. With the advance of robotic simulation, RL-based locomotion is often trained in virtual environments~\cite{kumar2021rma,makoviychuk2021isaac,rudin2022learning} with diverse terrains~\cite{rudin2022learning} and randomized environmental factors~\cite{lee2020learning,kumar2021rma} to improve the policy robustness. This technique is commonly known as domain randomization (DR). Training in simulators also provides rich information, some of which is not easily accessible in the real world (i.e., privileged information). To better interpret such information and bridge the sim-to-real gap when deploying policies to the physical world, system identification is commonly used to transfer knowledge to a deployable student policy.  For instance, Lee et al.~\cite{lee2020learning} employed action imitation to infer teacher behaviors using historical proprioceptive data. Kumar et al.~\cite{kumar2021rma} further developed a two-stage adaption framework for faster and more robust online transfer, which has become the foundation for many subsequent works~\cite{margolis2024rapid,lai2023sim}. Another approach combines transfer loss with RL loss for joint optimization~\cite{liu2022saving,radosavovic2023learning} to allow the student to explore with teacher guidance to the maximization of reward return. 

\subsection{Transformer in Robotics}
The transformer-based models have been introduced to solve robotic tasks. For instance, Decision Transformer~\cite{chen2021decision} converts states, actions, and rewards into embeddings using an encoder. Trajectory Transformer~\cite{janner2021offline} uses the complete discretized trajectory for language model-like autoregressive prediction. Building on these frameworks, Gato~\cite{reed2022generalist} was developed to serve as a general agent for hundreds of tasks, including real-world robotic manipulation. Vision Language Models (VLM)~\cite{brohan2022rt} and Vision Language Action Models (VLA)~\cite{brohan2023rt,padalkar2023open} use transformers as interfaces for scene and language understanding to provide high-level commands for robotic control and human-robot interaction but due to the enormous model size, they often run at only 5 Hz~\cite{brohan2022rt,brohan2023rt}.

For legged locomotion, which requires real-time control, some work involves outputting high-level commands as an interface. For instance, 
Yang et al.~\cite{yang2021learning} developed a transformer model for vision-based locomotion, which outputs high-level velocity commands and relies on a dedicated low-level controller for motor control. Similarly, Tang et al.~\cite{tang2023saytap} uses gait pattern as the interface for a low level controller. The external controller often requires additional design and training, and to bring transformers to direct motor control, Lai et al.~\cite{lai2023sim} proposed TERT, which utilizes historical observation-action pairs to generate target motor commands directly. Barkour~\cite{caluwaerts2023barkour} uses a similar architecture to merge multiple specialist policies into a single locomotion policy. Radosavovic et al.~\cite{radosavovic2023learning} applied a similar method to the bipedal locomotion task and later reformulated it as a next token prediction problem~\cite{radosavovic2024humanoid}. Recently, Sferrazza et al. proposed BoT~\cite{sferrazza2024body}, an embodiment-aware transformer network with body-induced bias based on the embodiment graph with embody-specific masking. However, to achieve real time onboard inference, end-to-end controllers relies on fixed information on each timestep and node for ecoding, exhibiting inflexibility of handling different inputs. 

\section{Preliminary}
We adopt the two-stage teacher-student transfer approach from TERT~\cite{lai2023sim} as the basis, which utilizes a well-trained teacher policy through RL with privileged information.

\subsection{Simulation Environment}

We implement the simulation environment based on Isaac Gym and its open-source library IsaacGymEnvs~\cite{makoviychuk2021isaac} to enable massive parallel training.

\noindent\textbf{Terrain and Curriculum.} We adopt the terrain curriculum from~\cite{rudin2022learning} with five terrain types (smooth slope, rough slope, stairs up, stairs down, discrete obstacle) and difficulty curriculum. The agent progresses and regresses the level based on the episode cumulative tracked linear reward.

\noindent\textbf{Domain Randomization.}
To enhance the robustness of the policy, DR is used in the simulation following~\cite{rudin2022learning,nahrendra2023dreamwaq}. We sample the commanded longitudinal and lateral velocity from [-1.0, 1.0] m/s, and horizontal angular velocity first calculated based on sampled heading and capped at [-1.0, 1.0] rad/s. Due to the significant computation required for transformer, a system delay is added~\cite{nahrendra2023dreamwaq}.

\noindent\textbf{Observations and Actions.} The privilege observation $e_t$ for teacher training contains ground-truth data gathered from simulation, including base linear and angular velocity, orientation, surrounding height map and randomized parameters as described above. For proprioceptive information, we use three commonly seen low-level sensors from quadrupeds. i.e., joint encoders, IMU and foot contact sensors. These sensors can provide five sensory data, including joint position $q \in \mathbb{R}^{12}$, joint velocity $\dot{q} \in \mathbb{R}^{12}$, angular velocity $\omega \in \mathbb{R}^{3}$, gravity vector $g \in \mathbb{R}^{3}$ and binary foot contact $c \in \mathbb{R}^{4}$. Furthermore, the randomly sampled user command target $cmd = [v_x, v_y, \omega_z]$ and actions from previous step $a_{t-1} \in \mathbb{R}^{12}$ are added, resulting in an observation of $o_t \in \mathbb{R}^{49}$ for each step. To gather the temporal information, a list of historical proprioceptive information $[o_0, o_1, \cdots, o_T]$ from past $T=15$ steps is stored. Thus, full observation is in the $\mathbb{R}^{49 \times 15}$ space. Both the teacher and student output the desired joint position $a_t$, which is further processed by a PD controller for the output torque $\tau = K_p (\hat{q}-q) + K_d (\hat{\dot q}- \dot q)$, with base stiffness and damping set to 30 and 0.7 respectively and the target joint velocity $\hat{\dot q}$ set to 0.

\noindent\textbf{Reward Function for RL.} The reward functions are designed to encourage the agent to follow the commanded velocity. Following~\cite{rudin2022learning,kumar2021rma,nahrendra2023dreamwaq}, we primarily penalize the linear and angular movement along  other axes, large joint acceleration and excessive power consumption.

\begin{figure*}[t]
    \centering
    \includegraphics[width=.85\textwidth]{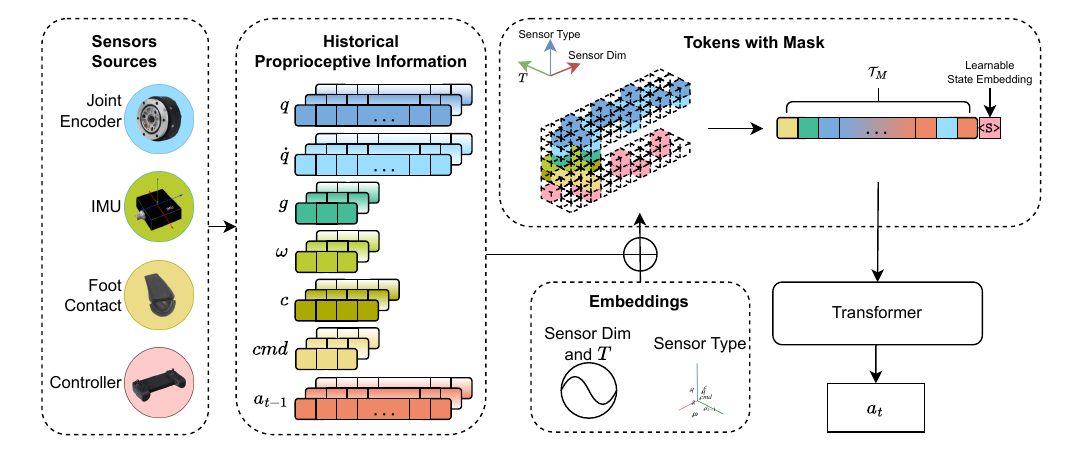}
    \caption{Overview of our \methodname{}. We gather proprioceptive information from commonly seen low-level sensors for discretization and tokenization. Similar to video understanding, we add additional embedding in three dimensions: sensor type, sensor dim and time. Before being passed to the transformer, a random mask is applied to partially remove the information and a learnable state embedding \texttt{<S>} is used to consolidate the information for action prediction. The target joint position output is passed to the PD controller for direct joint control.}
    \label{fig:overview}
\end{figure*}

\subsection{Teacher Policy and Training}

We implement a teacher policy following~\cite{kumar2021rma}. The teacher first encodes the privilege information $e_t$, with a factor encoder $\mu$ into a latent space $l_t$, which is then combined with the latest observation-action pair $o_t$ for the teacher policy $\hat{\pi}$ to output the desired joint position $\hat{a}_t$:
\begin{equation}
    \begin{gathered}
        \hat{l_t} = \mu(e_t), \;\;\; \hat{a}_t = \hat{\pi} (\hat{l_t}, o_t) \label{eq:teacher}
    \end{gathered}
\end{equation}

The $\mu$ and $\pi$ networks are implemented as MLP with hidden layers of $[512, 256, 128]$ and $[256, 128]$, respectively. The teacher policy is trained with PPO~\cite{schulman2017proximal} directly to maximize the reward return and is shared across all student transfers at later stages for a fair comparison.

\section{Methodology}

We present \methodname{}, a novel transformer-based model to generate a generalized understanding of low-level proprioceptive information for quadruped locomotion in complex terrains to handle different sensor set equipped on various robot models or when the sensors are damaged and not available. Unlike previous works~\cite{lai2023sim,radosavovic2023learning,radosavovic2024humanoid}, where each observation-action pair is processed at the timestep level, we treat each sensor modality individually so that the transformer can learn at the lowest level possible. With this foundational understanding, our model is capable of handling different combinations of sensor inputs, enabling better generalization and flexibility. It can potentially be extended to incorporate high-dimensional sensors for more complex tasks. Fig.~\ref{fig:overview} shows the overview of \methodname{}.

\noindent\textbf{Sensory-Action Data Tokenize.}
To learn in-context information at the lowest sensor level, each modality is encoded individually. Instead of the linear projection used in previous transformer-based locomotion controllers~\cite{lai2023sim,radosavovic2023learning,radosavovic2024humanoid}, where all sensory observations are merged, individual continuous sensor and control data are mapped to tokens directly. Following previous works~\cite{reed2022generalist, brohan2022rt, brohan2023rt}, we pass the normalized data through an encoder to discretizes the value into 256 bins, which are further mapped into a learnable embedding space with $d=128$ dimensions. Compared to timestep level encoding, in this way, the most information is preserved for in-context understanding by transformer.

\noindent\textbf{Positional Embedding and Sensor Type Embedding.} We view the encoded information in a three-dimensional way, sensor type, sensor channel and timestep. To distinguish proprioceptive information from different sources with temporal relations, two additional embeddings are added. The first one is a fixed 2D sin-cos position embedding $e_p$~\cite{beyer2022better} applied on the channel dimension and time axis of each sensor. For instance, $e_P^{i,t}$ means the embedding added to the \textit{i}-th channel at timestep $t$ This allows the model to handle sensors with varying lengths of dimensions and historical time windows directly and be easily extendable. To accommodate the multimodal nature of the sensory data, another learnable embedding $e_S$ is add to indicate each sensor type. This enables easy mix and match of information from different sensors without concerns about the order or placeholders. When new sensors are added, a new sensor embedding can be trained and added in. Thus, for the embedding of \textit{i}-th channel of \textit{a} sensor at timestep $t$, with original encoded token embedding $e_t^{a,i}$, the finial value in the sequence $\mathcal{T}$ is:

\begin{equation}
    \mathcal{T}_t^{a,i} = e_t^{a,i} + e_P^{i,t} + e_S^a
\end{equation}

\noindent\textbf{Random Masking.} Inspired by the use of masking in image and video understanding~\cite{he2022masked,feichtenhofer2022masked} with autoencoders to improve vision understanding, we create a binary mask $M$ based on the target ratio $\alpha$ to randomly mask out portions of the collected sensory, which are directly removed tokens from the original sequence:
\begin{equation}
    \mathcal{T}_M = \{e_i \in \mathcal{T}: M_i = 1\}
\end{equation}

Since only part of the sensory data are visible to the network, the model is required to infer and reconstruct the missing information from them, thereby enhancing its understanding of the relationships between different sensory inputs. Furthermore, random masking significantly reduces the training time and computational resources required. With sensor level tokens, the input sequence length grows from $T$ to $49T$ for observation, and as the complexity of self-attention is necessarily quadratic in the input length~\cite{keles2023computational}, the added overhead is enormous, and masking makes it more feasible to run during massive parallel training.

\noindent\textbf{Transformer Model.} We implement a vanilla transformer model to process the generated tokens. The model consists of multihead self-attention blocks with an MLP ratio of 2.0. An additional learnable state embedding \texttt{<S>} is added to the end of the masked sequence $\mathcal{T}_M$ to consolidate the processed information~\cite{dosovitskiy2020image}, which is subsequently projected into the action space with an MLP network $\pi$:
\begin{equation}
    \begin{gathered}
        l_t = \mathrm{\methodname{}}([\mathcal{T}_M, \texttt{<S>}]), \;\;\;\; a_t = \pi (l_t) \label{eq:transformer}
    \end{gathered}
\end{equation}

\noindent\textbf{Teacher-Student Transfer.} Following TERT~\cite{lai2023sim}, we train \methodname{} with a two-stage transfer strategy. In the first offline pretraining stage, trajectory is gathered by unrolling the well-trained teacher policy while the student will predict the next actions. This is to ensure that the student can produce reasonable actions during the second online correction stage to overcome the gap of distribution shift by training on its own trajectory. We minimize the loss for action prediction:
\begin{equation}
    \mathcal{L} = \| a_t - \hat{a}_t \| ^2
\end{equation}

\section{Experiments and Results}

We design and conduct various simulation experiments to evaluate the effectiveness of the proposed \methodname{}, and its generalization ability for different sensor data. We mainly adopt three metrics: linear velocity tracking return per step, angular velocity tracking return per step, and total final reward return. They indicate how the agent can conduct the task following users' commands and the overall performance. All reported results are averaged over 5000 trails with five terrain types and different levels. They are normalized on the basis of respect teacher data for easy comparison. 

\subsection{Impact of Mask Ratio}
\label{sec:mask}
\begin{figure}[t]
    \centering
    \includegraphics[width=0.9\columnwidth]{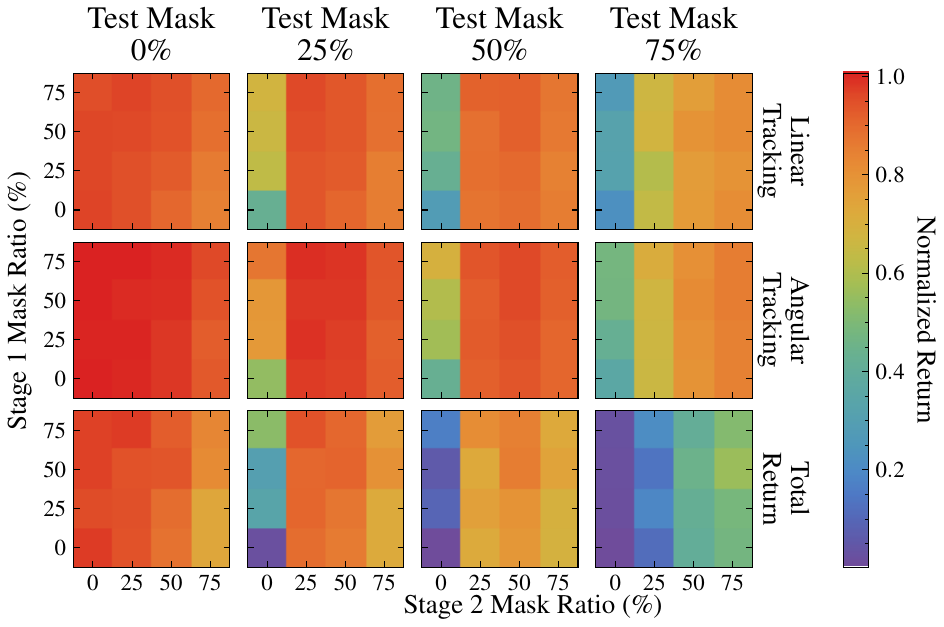}
    \caption{Heatmap matrix for the performance of models that are trained with different combinations of mask ratios. The three rows from top to bottom represent the linear velocity tracking, angular velocity tracking and total reward return respectively. The four columns denote different masking ratios applied during testing. For each sub-figure, the y-axis is the masking ratio applied during the offline pretraining stage and the x-axis is the masking ratio applied during the online correction stage.}
    \label{fig:mask_matrix}
\end{figure}

First, we investigate the maximum portion of missing data that \methodname{} can handle to reconstruct robot states. During each the transfer stages, we set the masking ratio to 0\%, 25\%, 50\% and 75\% independently. Fig.~\ref{fig:mask_matrix} shows the resultant heatmap matrix. When trained without masking, despite the model having very good performance with all the information available, it suffers from missing data and cannot efficiently reconstruct the status. We can also see that the performance is more dependent on the masking ratio in the second stage than that in the first stage. This is because in the second transfer stage, the student is interacting with the environment to reduce the gap caused by missing information and observation shift. In contrast, the mission of the first stage is to generate a usable policy that outputs reasonable actions so the agent does not fail dramatically and has the chance in the second stage to generate high-quality trajectories for optimization, which is achievable even with a masking ratio of 75\%. This also demonstrates the importance of using two-stage transfer. 

Comparing the performance of these models, we choose the one trained with the masking ratio of 75\% in the first stage and 50\% in the second stage, which can well balance the resource requirement and agent performance.

\begin{table*}[t]
\centering
\caption{Comparison results on different terrain types in terms of linear velocity tracking, angular velocity tracking and total reward return for all the variations of trained models.}

\label{tab:comparison}
\resizebox{\textwidth}{!}{
\begin{tabular}{cccccccccccccccccccc}
\Xhline{1.5pt}
\multirow{2}{*}{\textbf{Terrain}}                                       & \multirow{2}{*}{\textbf{Metric}} & \multicolumn{3}{c}{\textbf{Ours}}            & \multirow{2}{*}{\textbf{RMA}} & \multicolumn{2}{c}{\textbf{TERT}}   &  & \multicolumn{3}{c}{\textbf{GRU}}             &  & \multicolumn{3}{c}{\textbf{TERT w/ Mask}}    &  & \multicolumn{3}{c}{\textbf{Ours w/ Learnable}} \\ \cline{3-5} \cline{7-8} \cline{10-12} \cline{14-16} \cline{18-20} 
                                                                        &                                  & \textbf{0\%} & \textbf{25\%} & \textbf{50\%} &                               & \textbf{Concat} & \textbf{Seperate} &  & \textbf{0\%} & \textbf{25\%} & \textbf{50\%} &  & \textbf{0\%} & \textbf{25\%} & \textbf{50\%} &  & \textbf{0\%}  & \textbf{25\%}  & \textbf{50\%} \\ \Xhline{1.5pt}
\multirow{3}{*}{\begin{tabular}[c]{@{}c@{}}Smooth\\ Slope\end{tabular}} & Linear Tracking                  & 1.00         & 0.99          & 0.99          & 1.00                          & 1.01            & 1.00              &  & 1.00         & 0.96          & 0.78          &  & 0.95         & 0.98          & 0.99          &  & 0.21          & 0.93           & 0.99          \\
                                                                        & Angular Tracking                 & 1.02         & 1.02          & 1.00          & 1.02                          & 1.02            & 1.02              &  & 1.02         & 1.01          & 0.97          &  & 0.99         & 1.00          & 0.99          &  & 0.35          & 0.97           & 0.99          \\
                                                                        & Total Reward                     & 1.01         & 1.01          & 0.99          & 1.02                          & 1.02            & 1.02              &  & 1.01         & 0.99          & 0.84          &  & 0.95         & 0.99          & 0.98          &  & 0.25          & 0.92           & 0.99          \\ \hline
\multirow{3}{*}{\begin{tabular}[c]{@{}c@{}}Rough\\ Slope\end{tabular}}  & Linear Tracking                  & 0.97         & 0.94          & 0.95          & 0.99                          & 1.00            & 0.98              &  & 0.96         & 0.90          & 0.74          &  & 0.91         & 0.95          & 0.93          &  & 0.21          & 0.87           & 0.95          \\
                                                                        & Angular Tracking                 & 1.01         & 1.00          & 0.98          & 1.02                          & 1.02            & 1.01              &  & 1.01         & 1.00          & 0.95          &  & 0.96         & 0.97          & 0.94          &  & 0.33          & 0.94           & 0.96          \\
                                                                        & Total Reward                     & 0.98         & 0.95          & 0.95          & 1.00                          & 1.01            & 0.99              &  & 0.97         & 0.92          & 0.79          &  & 0.90         & 0.94          & 0.89          &  & 0.19          & 0.80           & 0.91          \\ \hline
\multirow{3}{*}{\begin{tabular}[c]{@{}c@{}}Stairs\\ Up\end{tabular}}    & Linear Tracking                  & 0.93         & 0.91          & 0.91          & 0.94                          & 0.95            & 0.95              &  & 0.90         & 0.91          & 0.69          &  & 0.82         & 0.86          & 0.85          &  & 0.26          & 0.81           & 0.85          \\
                                                                        & Angular Tracking                 & 1.00         & 0.98          & 0.97          & 0.99                          & 1.00            & 1.01              &  & 0.99         & 0.98          & 0.94          &  & 0.95         & 0.95          & 0.92          &  & 0.71          & 0.94           & 0.93          \\
                                                                        & Total Reward                     & 0.95         & 0.91          & 0.87          & 0.95                          & 0.99            & 1.00              &  & 0.90         & 0.87          & 0.72          &  & 0.80         & 0.81          & 0.73          &  & 0.54          & 0.74           & 0.72          \\ \hline
\multirow{3}{*}{\begin{tabular}[c]{@{}c@{}}Stairs\\ Down\end{tabular}}  & Linear Tracking                  & 0.94         & 0.93          & 0.93          & 0.96                          & 0.97            & 0.95              &  & 0.92         & 0.93          & 0.74          &  & 0.86         & 0.91          & 0.92          &  & 0.71          & 0.94           & 0.93          \\
                                                                        & Angular Tracking                 & 1.00         & 0.99          & 0.99          & 1.00                          & 1.01            & 1.01              &  & 0.99         & 0.98          & 0.95          &  & 0.95         & 0.96          & 0.92          &  & 0.70          & 0.93           & 0.93          \\
                                                                        & Total Reward                     & 0.94         & 0.91          & 0.91          & 0.95                          & 1.00            & 0.98              &  & 0.89         & 0.90          & 0.77          &  & 0.83         & 0.86          & 0.77          &  & 0.49          & 0.75           & 0.74          \\ \hline
\multirow{3}{*}{Discrete}                                               & Linear Tracking                  & 0.96         & 0.96          & 0.94          & 0.97                          & 0.99            & 0.98              &  & 0.93         & 0.88          & 0.75          &  & 0.87         & 0.91          & 0.89          &  & 0.25          & 0.87           & 0.85          \\
                                                                        & Angular Tracking                 & 1.01         & 1.01          & 0.99          & 1.01                          & 1.02            & 1.02              &  & 1.01         & 0.99          & 0.95          &  & 0.96         & 0.97          & 0.94          &  & 0.37          & 0.94           & 0.94          \\
                                                                        & Total Reward                     & 1.00         & 0.97          & 0.96          & 0.99                          & 1.04            & 1.02              &  & 0.94         & 0.89          & 0.83          &  & 0.88         & 0.91          & 0.80          &  & 0.03          & 0.80           & 0.76          \\ \Xhline{1.5pt}
\multirow{3}{*}{\textbf{Average}}                                       & Linear Tracking                  & 0.96         & 0.95          & 0.94          & 0.97                          & 0.98            & 0.97              &  & 0.94         & 0.92          & 0.74          &  & 0.88         & 0.92          & 0.92          &  & 0.23          & 0.87           & 0.91          \\
                                                                        & Angular Tracking                 & 1.01         & 1.00          & 0.99          & 1.01                          & 1.01            & 1.01              &  & 1.00         & 0.99          & 0.95          &  & 0.96         & 0.97          & 0.94          &  & 0.49          & 0.95           & 0.95          \\
                                                                        & Total Reward                     & 0.97         & 0.95          & 0.94          & 0.98                          & 1.01            & 1.00              &  & 0.94         & 0.92          & 0.79          &  & 0.87         & 0.90          & 0.83          &  & 0.30          & 0.80           & 0.82          \\ \Xhline{1.5pt}
\end{tabular}
}
\end{table*}

\subsection{Comparison with Baselines}

We compare \methodname{} with two baselines. The first is RMA \cite{kumar2021rma}, which is implemented with TCN~\cite{bai2018empirical} to capture temporal information. The second is TERT \cite{lai2023sim}, a transformer-based framework with linear projection for observations and actions in two favors: concatenated single token and separate tokens for states and action, resulting in $T$ and $2T$ tokens respectively~\cite{radosavovic2024humanoid}. To evaluate the masking mechanism in our method, we replace the selected observation in \methodname{} with a learnable representation instead of removing them. To further evaluate the importance and capability of the transformer structure, we replace it with a GRU~\cite{chung2014empirical} model. We expand the missing information testing to TERT. However, since the observations and actions in TERT are encoded through linear projection before passing to the transformer, it is impossible to directly remove any input. Thus, the the same learnable masks method is applied to TERT.

All variations of \methodname{}, TERT and other baselines are trained with the same two-stage transfer, sharing a common well-trained teacher network, and we apply a testing mask of up to 50\%, as identified in Section~\ref{sec:mask}. 

Tab.~\ref{tab:comparison} shows the comparison results. When fully optimized with teacher-student transfer, the performance of all fully trained vanilla policies with complete observations is very close, often within just 2\% difference. When faced with incomplete information, transformer-based \methodname{} can have a better understanding of the data and reconstruct the robot state more accurately than the GRU-based network, even with only half of the information. When using a learnable representation mask, with \methodname{} or TERT, the agent underperforms to the vanilla removing mask, especially with full observation, showing that a direct removing mask has an advance in both better performance and less resource required. Although we can hack the linear projection in the TERT network to take in missing information, it is not comparable to direct sensor level tokenzation and attention for sensory information understanding.

\subsection{Generalization, Robustness and Flexibility}

While achieving state-of-the-art performance, \methodname{} offers additional benefits of generalization and flexibility to customize the model after training or even on the fly to fit the deployment requirement. Quadrupeds are equipped with different sensor sets, and sensor damage can cause certain channels or the entire sensor to be unavailable during deployment, which required the robustness against missing information to handle. Furthermore, we can balance the performance and required computation power by using a shorter sequence based on the insights from in-context sensory information understanding.

\begin{figure}[t]
    \centering
    \includegraphics[width=0.9\columnwidth]{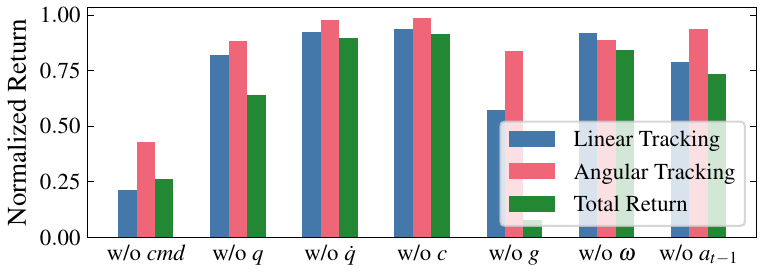}
    \caption{Performance with certain sensory feedback completely removed.}
    \label{fig:sensor_importance}
\end{figure}

\begin{figure}[t]
    \centering
    \includegraphics[width=0.9\columnwidth]{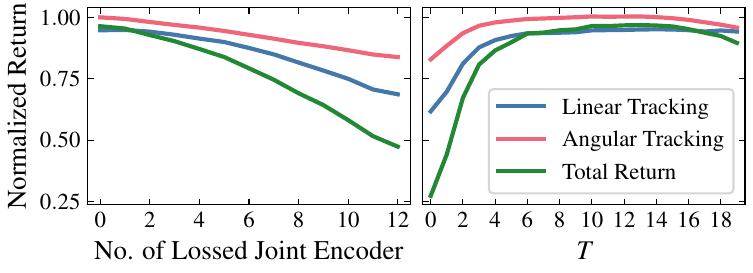}
    \caption{Performance with various setups:  \textbf{Left} certain numbers of joint encoders are masked out;\label{fig:dof_mask} \textbf{Right} different history time window $T$ is applied.\label{fig:time}}
    \label{fig:experiments}
\end{figure}

\noindent\textbf{Important Sensory Feedback.}
To understand the importance of each sensory feedback, we further investigate the impact of removing each sensor completely from the observation and the results are shown in Fig.~\ref{fig:sensor_importance}. It is clear that certain feedback like $\dot{q}$, $c$, $\omega$ and even $a_{t-1}$ are quite redundant and a well trained transformer-based \methodname{} can easily compensate the missing information from other sources, while the other sensor data are more critical.

\noindent\textbf{Missing of Sensor Dimension.} Some proprioceptive information has multiple channels, such as joint encoders and force senors. This means that these sensors can also be damaged independently due to wear and tear from daily operations and it is not easy to have a redundant sensor. Among these sensors, joint encoders are the source of both $q$ and $\dot{q}$ for the observation. From previous analysis, missing of joint information can be crucial. We investigate the scenario where only a few encoders are dead or the data are compromised and need to be excluded. We conduct the test by masking certain numbers of joint encoders and for each masked joint, the related $q$ and $\dot{q}$ are removed completely from the observation. The results are shown in Fig.~\ref{fig:dof_mask}. The loss of the joint encoder can have a great impact on the performance as the related information is very essential for quadruped locomotion. However, our transformer model can still handle multiple missing encoders before large performance degradation.

\noindent\textbf{Time Window.} \label{sec:time}
Another special masking is to completely remove some timesteps, the default window, $T=15$, is equivalent to past 0.3s. We check whether such a long sequence of information is necessary by applying different time windows without masking. The results are shown in Fig.~\ref{fig:time}. It is clear that \methodname{} can efficiently extract and reconstruct the robot state for actions even with only 7 steps of past information. Interestingly, given a longer timeframe like $T=20$, which the transformer has never seen during training, \methodname{} is still robust and not affected by such unknown information.

\begin{figure}[t]
    \centering
    \includegraphics[width=0.9\columnwidth]{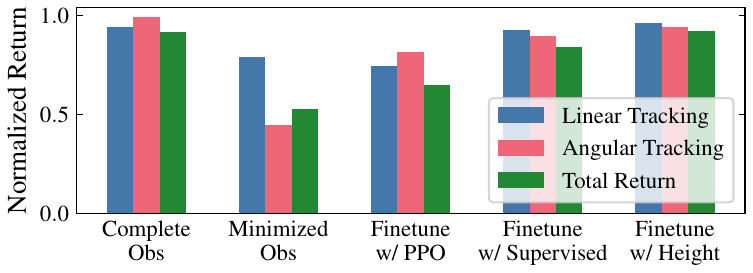}
    \caption{Performance using minimized observations with finetuning and extension of height map.}
    \label{fig:finetune}
    \vspace{-10pt}
\end{figure}

\noindent\textbf{Minimized Observation and Fine-tuning.} It is not feasible to infer the transformer with full observation space and long windows with the limited computation power onboard. From the previous analysis, we have identified the important sensors and the minimal history length required. We further explore the feasibility of creating a minimized observation policy based on the information. Using an observation with only $cmd$, $q$, $g$ and $a_{t-1}$ with a window of $T=7$, we are essentially removing 71\% of the tokens from the complete training observation space. 

When directly deployed with such mask, the policy cannot perform well due to all the missing information. To restore the performance, we freeze the transformer for fast fine-tuning of the projection layers and test both the vanilla PPO~\cite{schulman2017proximal} and supervised learning with online correction~\cite{lai2023sim}. The performance of the policies is shown in Fig.~\ref{fig:finetune}. While both algorithms can help improve the performance of the policy with only minimized observations, supervised learning gives larger boost. Training with the teacher has been identified as one major approach to achieve quadruped locomotion on challenging terrains~\cite{kumar2021rma,lai2023sim}. Although our foundation with the transformer can provide a solid start point of student policy, additional work is still needed for pure RL-based fine-tuning to reduce the dependence on privilege information. 

\noindent\textbf{Extension with New Information.} In previous analysis, \methodname{} is robustness against new timestep information. When extending the capability for quadrupeds, additional sensors such as cameras and LiDAR are often needed. We assess the model's capability of handling previously unseen information, which can be appended into $\mathcal{T}$ as new tokens. For instance, we tokenize the height map information using a vanilla MLP encoder and directly extend it to our minimized observation agent for fine-tuning. The performance of the extended agent is shown in Fig.~\ref{fig:finetune}. Height information significantly aids in navigating challenging terrains, such as staircases, and improves the overall locomotion performance even with minimal observations and new encoder needed to be trained. To take the test to an extreme, we added 256 randomly generated dummy tokens, equivalent to a camera frame with ViT~\cite{dosovitskiy2020image} before processing, and the agent can still produce explore, which is crucial for two-stage knowledge transfer. This demonstrates that the model can be used as a solid foundation for further extension with high-dimensional information by direct deployment in virtual environments to gather new trajectories. Please refer to the supplementary video for more information.

\begin{figure}[t]
    \centering
    \includegraphics[width=.95\columnwidth]{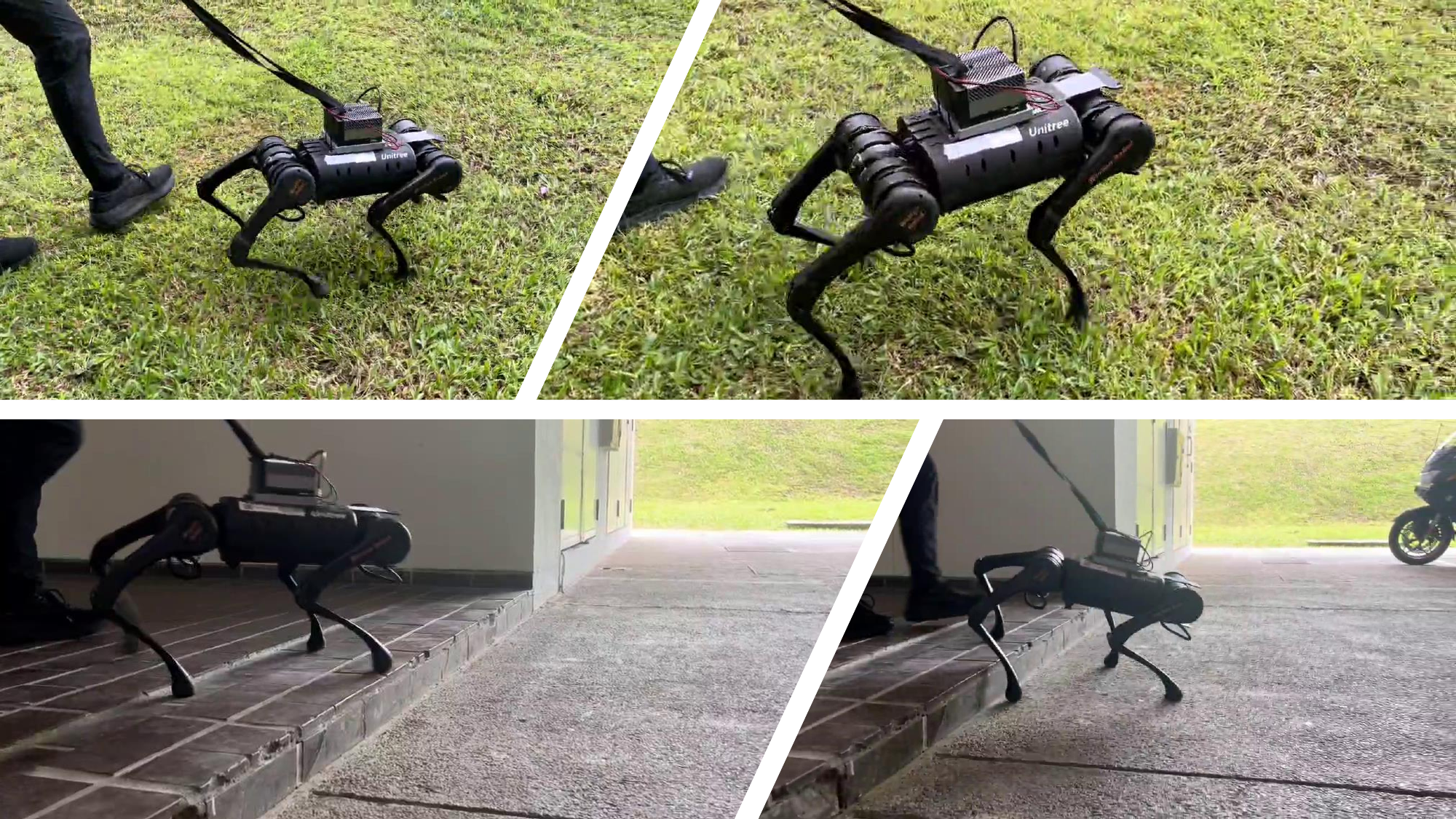}
    \caption{Deployment in the physical world on Unitree A1 with minimized observations with zero-shot transfer.}
    \label{fig:deployemnt}
\end{figure}

\subsection{Physical Deployment}
We successfully deploy the trained policy, exported with JIT, directly on a Unitree A1 robot equipped with a Jetson AGX Orin Developer Kit. The Jetson acts as both the main processor and a payload. No further model optimization is required for a zero-shot transfer. With the onboard processing power, the policy can run at 150Hz with our minimized observation, meeting the requirements for real-time deployment and allowing room for further extension with high-dimensional sensors. However, we notice that the JIT model will have slight difference in output compared to the original model, indicating additional work is needed for better portability. Fig.~\ref{fig:deployemnt} shows some snapshots from the deployment test. Please refer to the supplementary video for more information.

\section{Conclusion}

This paper introduces \methodname{}, a transformer-based model for quadruped locomotion. It leverages the masking technique and direct sensor-level attention to enhance the understanding and generation of sensory information input. We evaluate the robustness of \methodname{} with different combinations of proprioceptive information and demonstrate its capability to compensate for missing data and handle unseen information. Finally, we show that \methodname{} is efficient to be deployed on a physical robot without any additional optimization. 

Attention in the full sensory-temporal observation space is computationally intensive and time-consuming. Although using masking can significantly reduce the resources needed, it still takes hours for knowledge transfer, which is considerably longer than existing methods like TCN and temporal-level attention. Additionally, fine-tuning the model with pure reinforcement learning remains challenging, necessitating a more efficient knowledge transfer solution to leverage privileged information effectively. While the policy demonstrated its capability to handle missing data, an addition module is needed to detect and mask out the defected sensors. These will be our future work. 

\clearpage
\bibliographystyle{IEEEtran}
\bibliography{references.bib}

\end{document}